\documentclass[twocolumn]{article} 
\usepackage{cec2003,multicol,times} 
\usepackage{epsfig} 
\usepackage{amsmath}
\newcommand{\commentaire}[1]{ } 
\begin{document} 
\pagestyle{empty}
\sloppy
\twocolumn[ 
\title{Where are Bottlenecks in NK Fitness Landscapes?}
\vspace{0.1in}

\begin{center}
\textbf{S\'ebastien Verel, Philippe Collard and Manuel Clergue}  \\
Universit\'{e} de Nice-Sophia Antipolis, \\
06640, Laboratoire I3S, France, \\
\{verel,pc,clerguem\}@i3s.unice.fr\\
\end{center}
\vspace{0.1in}
]

\begin{abstract} 
Usually the offspring-parent fitness correlation is used to visualize and analyze some caracteristics of fitness landscapes such as evolvability. In this paper, we introduce a more general representation of this correlation, the \textit{Fitness Cloud (FC)}. We use the bottleneck metaphor to emphasise fitness levels in landscape that cause local search process to slow down. For a local search heuristic such as hill-climbing or simulated annealing, \textit{FC} allows to visualize bottleneck and neutrality of landscapes. To confirm the relevance of the \textit{FC} representation we show where the bottlenecks are in the well-know NK fitness landscape and also how to use neutrality information from the \textit{FC} to combine some neutral operator with local search heuristic.

\commentaire{
In this work the offspring-parent fitness correlation is used to visualize and analyze search spaces. We focus on the behavior of local search heuristics, as hill climbing or simulated annealing, on the well-known NK fitness landscape. We use the bottleneck metaphor to emphasise fitness levels in landscape that cause local search process to slow down. The \textit{fitness cloud} is used to identify a bottleneck's location at fitness level. Then, modeling dynamics by the way of the \textit{limit fitness cloud}, you are able to confirm the realness of bottlenecks in NK-landscape. Finally, the potential interest in combining local search techniques with neutral exploration is explored. }
\end{abstract}

\section*{Introduction} 
The fitness correlation between parent and offspring is often used to analyse search space. In this paper, we present the \textit{Fitness Cloud (FC)} which is the scatterplot of points parent-fitness/offpring-fitness. The $FC$ allowed us to visualize and analyse the dynamic of local search heuristic at fitness level. $FC$ shows evolvability as well neutrality and fitness bottleneck. The bottleneck value is the fitness level that cause local search to slow down and stop. In other word, bottleneck is fitness value where heuristic converge. In this paper, we focus on the $NK$ fitness landscapes. First, we present $NK$ fitness landscapes and the definition of Fitness Cloud. Section~2 reveals how $FC$ exhibit bottleneck in $NK$ landscapes for two well-know search heuristics hill climber and simulated annealing. The $FC$ represents the neutrality of landscape, section~3 proposes to use this information to design and analyse performances of strategy using neutrality.
\commentaire{
The paper is organized as follows. First, the basic features of Kauffman's NK fitness landscapes is briefly presented, then the evolvability on these landscapes is studied via the offspring-parent fitness correlation cloud. This \textit{fitness cloud} shape related to Hamming neighbourhood is analyzed. Section~2, presents the main contribution of this work; it reveals how fitness cloud takes heuristics into account and helps you to visualize and understand their dynamics. In this study we look for the average behavior only. According to the heuristic used, we show that it is possible to exhibit fitness bottlenecks. The usefulness of the proposed representation is demonstrated by experimental results obtained on two well-know search heuristics: \textit{hill climbers}, and \textit{simulated annealing}. Finally we tie together the results of previous section and suggest directions for further research. 
}

\section{NK Landscapes and Fitness Cloud}

\subsection{The Tunable NK-Fitness Landscapes} 
In this section the basic features of the family of NK fitness landscapes are reviewed. The notion of fitness landscapes \cite{WRI:69} as search space is defined as follows: a set of potential solutions (genotypes), a fitness function that evaluates the genotypes and a topology that represents relations between genotypes. NK model proposed by Kauffman \cite{KAU:93} is designed to capture the structure of rugged multi-peaked fitness landscapes. These random landscapes are defined on binary strings of length $N$. The parameter $K$ represents the number of \textit{epistatic} links\footnote{Epistasis is defined as the influence of the genotype at one locus on the effect of a mutation at another locus}. By tuning $K$, landscapes can generated with varying degrees of ruggedness. In order to compute the overall fitness of one string, one consider that each bit contributes a component to the total fitness based on its own value and the values of $K$ other genes. The random model is used, where fitness contribution of one bit depends on its own value and $K$ other randomly chosen bits\footnote{Weinberger \cite{WEI:96} proved the NK optimization problem with random neighbourhoods is \textsl{NP} complete for $K \ge 3$}. Fitness contributions come from a uniform distribution ranging from $0.0$ to $1.0$. Fitness of a string is computed as the sum of bits contribution at all $N$ loci divided by $N$ for normalization to the range $[0;1]$. 
The case $K=0$ corresponds to problem without epistasis: fitnesses of neighbourhood points are correlated. There exists a single optimum. A hill climbing search allows to reach this optimum and adaptive walks are thus relatively long (${N \mathord{\left/{\vphantom {N 2}} \right. \kern-\nulldelimiterspace} 2}$). The case $K=N-1$ corresponds to the maximum number of interaction of the parts. The fitness of any point is random. There exists an enormous number of local optima. Adaptive walks are relatively short ($\log(2N)$) and are likely to end up in local optimum.

\subsection{The Offspring-Parent Fitness Correlation Cloud} 
Plotting fitness against some features is not a new idea. B. Manderick et al. \cite{MAN:91} study the correlation coefficient of genetic operators: they compute the correlation between the fitnesses of a number of parents and the fitnesses of their offspring. J. Grefenstette \cite{GRE:95} uses fitness distribution of genetic operators to predict GA behaviour. H. Ros\'e et al. \cite{ROS:96} develop the \textit{density of states} approach by plotting the number of genotypes with a same fitness value. Smith et al. \cite{SMI:01} focus on notions of \textit{evolvability} and \textit{neutrality}; they plot the average fitness of offspring over fitness according to Hamming neighbourhood. \textit{Evolvability}  refers to the efficiency of evolutionary search. It is defined by Altenberg as "the ability of an operator/representation scheme to produce offspring that are fitter than their parents" (\cite{ALT:94},~\cite{WA-AL}).
\paragraph{Fitness Cloud}
In order to get a visual rendering of evolvability, we proposed a more general representaion in the plan parent-fitness / offspring-fitness. We consider that two strings are neighbours if there is a transformation related to a local search heuristics or an operator, which allows "to pass" from one string to the other one. For each string $x$ in the genotype space\footnote{Data are collected from an exhaustive enumeration of the search space}, we plot one point which have for abscissa the fitness $f(x)$ of $x$ and for ordinate the fitness $\tilde{f}(x)$ of a peculiar neighbour of $x$. Thus, we obtain a scatterplot, the so-called \textit{offspring-parent Fitness Cloud} ($FC$). The choice of one peculiar neighbour among all the possible ones is a feature of the heuristic. Implicitly the fitness cloud gives some insight on the genotype to phenotype map. The set of genotypes that all have equal fitness is a \textit{neutral set} \cite{KIM:83}. Such a set corresponds to one abscissa in the $FC$; according to this abscissa, a vertical slice from the cloud represents the fitness values that could be reached from this set of neutrality. For a given offspring-fitness value $\tilde f$, an horizontal slice represents all the fitness values from which a local operator can reach $\tilde f$. Evolvability against a fitness level can be charaterized by the repartition of points over the diagonal line in the $FC$. In this paper we will use the $FC$ to track the dynamic and to locate the bottlenecks of local search heuristic.
To get a more synthetic view on the $FC$, we define three functions :
$$ \tilde f_{min}(\varphi) = \underset{x \in G_{\varphi}}{min} \tilde f(x) $$
$$ \tilde f_{max}(\varphi) = \underset{x \in G_{\varphi}}{max} \tilde f(x) $$
$$ \tilde f_{mean}(\varphi) = \underset{x \in G_{\varphi}}{mean} \tilde f(x)$$

\begin{figure}[!h] 
\begin{center}  
\includegraphics[width=0.35\textwidth]{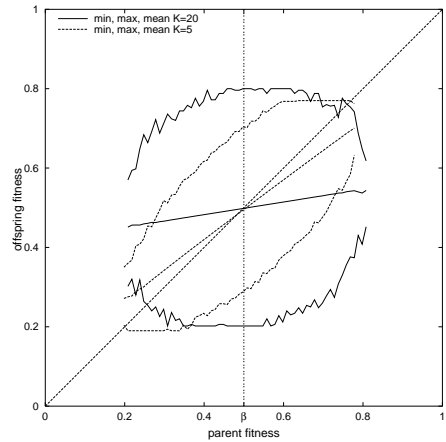} 
\caption{Fitness Cloud from the Hamming neighbourhood: border (min and max) and mean (NK-landscapes with $N=25$, $K=20$ and $K=5$)} 
\label{NK_25_20_5_contourHamming} 
\end{center}
\end{figure}

\commentaire{
the three following subsets are plotted:
\begin{center} 
$FC_{\min} = \left\{ {\left( {\varphi ,\tilde \varphi } \right)\ |\ \varphi \in f\left( {Gtype} \right),\ \tilde \varphi = \mathop {\min }\limits_{x \in G_\varphi } \tilde f\left( x \right)} \right\}$ 
\end{center}
\begin{center} 
$FC_{\max} = \{ {({\varphi ,\tilde \varphi })\ |\ \varphi \in f( {Gtype}),\ \tilde \varphi = \mathop {\max }\limits_{x \in G_\varphi } \tilde f(x)}\}$ 
\end{center}
\begin{center} 
$FC_{mean} = \left\{ {\left( {\varphi ,\tilde \varphi } \right)\ |\ \varphi \in f\left( {Gtype} \right),\ \tilde \varphi = \mathop {mean}\limits_{x \in G_\varphi } \tilde f\left( x \right)} \right\}$ 
\end{center}
}

where $G_\varphi$ is the neutral set defined by: $\left\{ {x \in Gtype\,\left| {\,f\left( x \right) = \varphi } \right.} \right\}$. Practically two fitness values are taken as equal if they both stand in the same interval\footnote{in our experiments the range is $0.002$}. We call $FC_{min}$, $FC_{max}$ and $FC_{mean}$ respectively the representative curve of $\tilde f_{min}$, $\tilde f_{max}$ and $\tilde f_{mean}$. Plotting the curves $FC_{max}$ and $FC_{min}$ allows to materialize the edge of the cloud (see fig. \ref{NK_25_20_5_contourHamming}). A peculiar fitness value $\beta$ is defined as solution of equation ${\tilde f_{mean}}
(\beta)=\beta$; it corresponds to the abscissa of intersection between the $FC_{mean}$ curve and the diagonal line (see fig. \ref{NK_25_20_5_contourHamming}). On average, $G_{\beta}$ is invariant by the heuristic, i.e. the heuristic is neutral on the set $G_{\beta}$.
\commentaire{On average a given heuristic is neutral on the set $G_{\beta}$.}

\paragraph{Offspring-Parent Fitness Cloud and Hamming Distance}
Before study a specific heuristic, it may be useful to get a view on the scatterplot fitness vs. fitness based on Hamming neighbourhood which is independent from any heuristic. So, we plot a cloud where all of the genotypes that can be produced by a single bit flip are selected (see fig.~\ref{NK_25_20_5_contourHamming}). Hence, the entire neighbourhood is represented without condition induced by some heuristics.

\begin{figure}[!h] 
\begin{center}  
\includegraphics[width=0.50\textwidth]{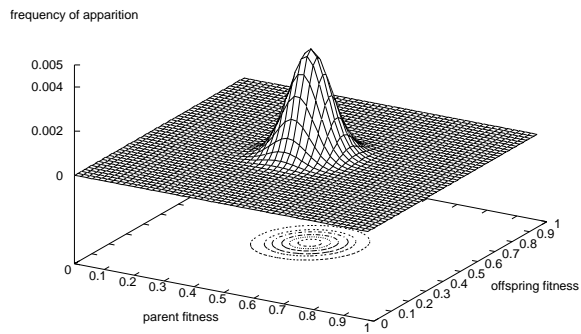}
\caption{Distribution of points on the Fitness Cloud (NK landscape with $N=25$ and $K=20$)}
\end{center}
\end{figure}

Figure \ref{NK_25_20_5_contourHamming} shows the $FC_{mean}$ curve is coarsely a line. This illustrates a well-known result: Weinberger \cite{WEI:90} established the following relation between the mean offspring-fitness and the length $N$, the epistatic parameter $K$ and the fitness value $f$: 
\begin{equation} 
\tilde f_{mean}(f) = \left( {1 - \frac{{K + 1}}{N}} \right) f + \left( {\frac{{K + 1}}{N}} \right)\beta 
\label{mean_fitness_line_hamming} 
\end{equation} 
where $\beta$ is constant. Therefore the mean offspring-fitness depends linearly from the parent-fitness whatever the epistatic parameter $K$ is. As reported by Smith \cite{SMI:01}, let's note that the slop coefficient ${1-\frac{{K + 1}}{N}}$ is the offspring-parent fitnesses correlation \cite{WEI:90}. The $\beta$ fitness level is always equal to $0.5$. So, when the parameter $K$ varies from $0$ to $N-1$, the $FC_{mean}$ line turns around the $({\beta ;\beta })$ point (see fig.~\ref{NK_25_20_5_contourHamming}). For $K=0$ the problem is linear and the $FC_{mean}$ line is near the diagonal; at the opposite when epistasis is upper limit ($K=N-1$), the $FC_{mean}$ line is close to the horizontal.

\section{Fitness Bottleneck and Limit Fitness Cloud}
In this section we show that the fitness cloud is useful in identifying a bottleneck's location at fitness level. In a first step we plot the $FC$ according to a local search heuristic. So we are able to locate the bottleneck. Then modeling dynamic by the way of the \textit{limit fitness cloud} allows to confirm the realness of bottleneck value.
In the following two computational search techniques are used: \textit{myopic hill climbing} (mHC) and \textit{simulated annealing} (SA). They implement adaptive local search; the neighbourhood is defined in terms of applying Hamming mutation.

\subsection{Modeling Dynamics at Fitness Level} 
Let H an heuristic, we assume there is a function $H(X)$ which allows to model the average dynamics of $H$ at fitness level. Given an initial genotype of fitness $f_1$, applying the heuristic generates a sequence $f_1,f_2,...$ by the iteration $f_{k + 1} = H( {f_k})$. Our hope is to gain knowledge from function $H$ in order to help us to predict the behaviour of the heuristic. To illustrate this approach, let us consider the heuristic (noted $H_{ham}$) corresponding to a \textit{random walk}: starting from a initial genotype, at each step the next genotype is chosen at random in the Hamming neighbourhood. Equation \ref{mean_fitness_line_hamming} may be reformulated in 
\begin{equation} 
H_{ham}(f) = \left( {1 - \frac{{K + 1}}{N}} \right) f + \left( {\frac{{K + 1}}{N}} \right)\beta 
\label{mean_fitness_line_hamming2} 
\end{equation}
From an initial fitness value $f_1$ less than $\beta$, the sequence $f_1,f_2,...$ increases to $\beta$. On average offspring-fitness is higher than parent-fitness; thus the heuristic is selectively advantageous. If fitness is greater than $\beta$, the mean offspring-fitness is lower than fitness: on average the heuristic is deleterious. Property \{$\beta=0.5$\} means that on average ${H}_{ham}$ is selectively neutral\footnote{${H}_{ham}$ induces no effect on fitness level} on NK-landscapes whatever epistasis is. Starting from $f_1=0.5$\footnote{Fitness of a random initial genotype is on average closed to the mean fitness over the search space ($\bar f=0.5$)} the heuristic generates the sequence $f_1=0.5,f_2=0.5,...$. 
In order to get a visual rendering of the long term behavior of an heuristic, for each string in the genotype space a point is plotted; the abscissa of which is the fitness $f$ and the ordinate the fitness $f^*$ of a genotype reached after applying the heuristic a given number of times. Thus, a new scatterplot, the so-called \textit{Limit Fitness Cloud} (noted $FC^*$), is drawn. We define the following function:
$$f^*_{mean}(\varphi) = \underset{x \in G_{\varphi}}{mean} f^*(x)$$
and the fitness value $\beta^*$ as $f^*_{mean}(\beta)$\footnote{of course, in this definition we assume that $\beta$ exists}. We call $FC^*_{mean}$ the representative curve of $f^*_{mean}$.

\subsection{Myopic Hill Climber} 
\begin{figure}[!h] 
\begin{center}
\includegraphics[width=0.35\textwidth]{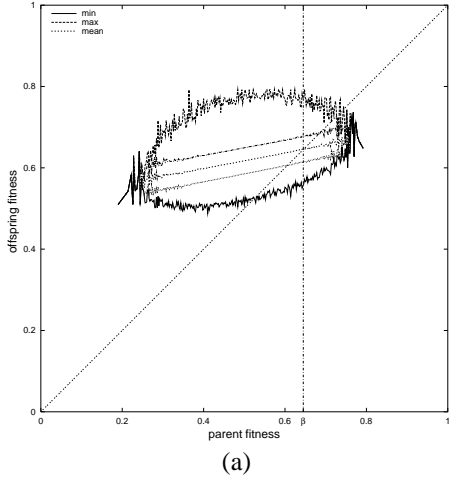} \\
(a) \\ 
\includegraphics[width=0.355\textwidth]{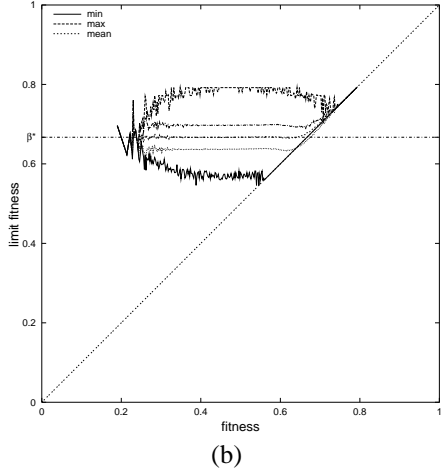} \\ 
(b) \\
\end{center}
\caption{Fitness Cloud under mHC: border (min and max) and mean (with standard deviation) ($FC$ (a), $FC^*$ (b)) (NK-landscape with $N=20$ and $K=15$)} 
\label{NK_20_15_best1} 
\end{figure}

A myopic hill climbing heuristic (so-called mHC) is used. At each step, the fittest of all of the genotypes that can be produced by a single bit flip is selected. Entire neighbourhood is searched and selection occurs in all cases, even when the best of the one-mutant neighbours of a genotype is less fit than it. 
Figure \ref{NK_20_15_best1} (a) shows the $FC_{mean}$ curve is coarsely a line too: the mean offspring-fitness is in proportion to fitness, whatever the epistatic parameter $K$ is. Relation between $\tilde f_{mean}$ and the length $N$, the epistatic parameter $K$ and fitness $f$ verifies the following equation:
\begin{equation} 
\tilde f_{mean}(f) = \left( {1 - \frac{{K + 1}}{N}} \right) f + \left( {\frac{{K + 1}}{N}} \right) E(X)
\label{mean_fitness_mHC} 
\end{equation}

The mean term $E(X)$ is equal to $E(X(N,K))$ where $X(N,K) = max( X_1, \ldots , X_N)$ and $X_i$ follows normal law ${\cal N} (0.5,\ \frac{1}{\sqrt{12(K+1)}})$. Then, correlation between parent-fitness and the mean offspring-fitness is linear. Therefore $\beta=E(X(N,K))$ is not any more constant but depends from $N$ and $K$. When $N$ is fixed, $\beta$ grows as the amount of epistasis decreases. A least squares regression is computed from the $FC_{mean}$ set for $N=20$ and $K=15$, we find:
\begin{equation} 
\tilde f_{mean} =  0.200 f + 0.516 
\label{mean_fitness_line_hamming_least_squares}
\end{equation}
which agrees to equation~\ref{mean_fitness_mHC}

The $FC$ allows to predict whether in the fitness space a bottleneck is likely to arise. We conjecture there is such a bottleneck in the vicinity of the $\beta$ fitness level. This means that, applying mHC heuristic from a point which fitness is below $\beta$, on average the search process breaks off around $\beta$. In particular from a random initial point, dynamic pushes fitnesses toward $\beta$. 
In order to validate this hypothesis, from each genotype as initial point, mHC is ran over 50 generations to collect the fitness of the last point encountered. All these informations are got together to build up the corresponding \textit{limit fitness cloud} see figure \ref{NK_20_15_best1} (b). According to the initial fitness value (abcissa) we can observe two main cases about the limit behavor of the mHC heuristic. Applying the heuristic a given number of times results to an equilibrium state where the mean fitness $f^*$ no longer changes from generation to generation. First, for an initial fitness below $\beta$, on average mHC "converges" to the fitness value $\beta^*$. Let's note that $\beta$ is smaller than $\beta^*$ but have the same magnitude (see tab.~\ref{table}). Second, fitnesses $f$ above $\beta$ are fixed points in the fitness space ($f^*=f$). Let's note that there is a transition range around $\beta$ where $f^*$ depends not linearly on $f$. These experiments support the \textit{bottleneck conjecture}: it is difficult to bypasse the $\beta^*$ fitness level for the mHC heuritic. Of course, breaking the bottleneck may occur for a particular initial genotype as in this study we look for the average behavior only.

\begin{table}[h]
\begin{center}
\begin{tabular}{|c|c|c|}
\hline
metaheuristic & $\beta$ &  $\beta^*$ \\
\hline
\hline
mHC & 0.645 & 0.667 \\
SA ($T=0.10$) & 0.524 & 0.559 \\
SA ($T=0.05$) & 0.548 & 0.590 \\
SA ($T=0.01$) & [0.604, 0.792]  & 0.656 \\
SA (Generation 50) & - & 0.560 \\
SA (Generation 1000) & - & 0.613 \\
SA (Generation 1900) & - & 0.682 \\
SA (Generation 2450) & - & 0.701 \\
nHC &  [0.686, 0.792]  & 0.746 \\
\hline
\end{tabular}
\end{center}
\caption{Experimental values of $\beta$ and $\beta^*$ for $N=20$ and $K=15$ with mHC, SA and nHC. The maximum fitness value for this fitness landscape is $0.792$}
\label{table}
\end{table}

\subsection{Simulated Annealing} 
Simulated annealing (SA) can be seen as a way of trying to allow solution to get away from local optima and move toward fitter point. The SA algorithm employs a random search that not only accepts changes that increase the fitness function, but also some changes that decrease the fitness value, thus allowing SA to jump out of local maxima. SA search technique takes its inspiration from the models of the annealing physical process \cite {KIR:83}. It is search process based on using a parameter which can play the role of temperature. The ability to avoid to get stuck in local optima depends on the choice of initial temperature, the number of iterations performed at each temperature, and the way the temperature is decremented. At each step, one genotype from all of the genotypes that can be produced by a single bit flip is selected and the resulting change, $\Delta f = \tilde f - f$, in fitness is computed. If $\Delta f > 0$, the new point is accepted; else, it is accepted with probability 
$e^{{{\Delta f} \mathord{\left/{\vphantom {{\Delta f} T}}\right.\kern -\nulldelimiterspace} T}}$, where $T$ is the temperature control parameter. One major problem with SA is to control the \textit{cooling process}. Often the cooling schedule is developed by trial and error for each particular landscape. First, with regard to given temperature values, the fitness cloud is analyzed; then a cooling process is implemented.
For a given temperature $T$, relation between $\tilde f_{mean}$, $f$, $N$, $K$ and $T$ can be derived from equation ~\ref{mean_fitness_mHC}. In this case, 
\begin{eqnarray*}
E(X) & = & E(X(f,N,K,T)) \\
& = & 1-\phi(\frac{f-0,5}{\sigma_K})+ \int_{- \infty }^{f}\varphi(\frac{x-f}{\sigma_K})e^{\frac{x-f}{T}}dx
\end{eqnarray*}
where $\varphi$ and $\Phi$ are respectively the density and repartition function of the reduced centered normal law. So $E(X)$ depends, not only on $N$ and $K$, but on the fitness value too. As a consequence, the set $FC_{mean}$ is no more represented by a line but by a curved shape.

\paragraph{High temperature}
\begin{figure}[!h] 
\begin{center} 
\includegraphics[width=0.35\textwidth]{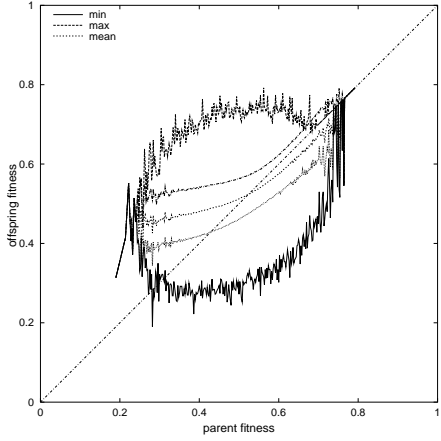} \\
(a) \\
\includegraphics[width=0.35\textwidth]{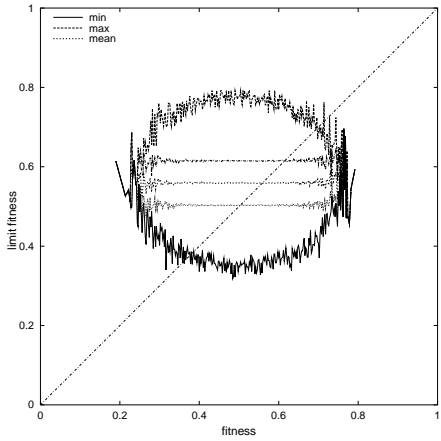} \\
(b) \\
\end{center} 
\caption{Fitness Cloud under SA at \textbf{high temperature}: border (min and max) and mean (with standard deviation) ($FC$ (a), $FC^*$ (b)) (NK-landscape with $N=20$ and $K=15$)} 
\label{NK_20_15_sa_0.10} 
\end{figure}

As predicted by our analytical study, at high temperature ($T=0.10$), the $FC_{mean}$ set is represented by a curved shape (see fig.~\ref{NK_20_15_sa_0.10} (a)). This curve crosses the diagonal at point $(\beta;\beta)$ distinctly; so it is easy to estimate $\beta$ (see tab.~\ref{table}). Plotting the limit fitness cloud $FC^*$ shows that it is difficult to bypasse the $\beta^*$ bottleneck level for the SA at high temperature (see fig.~\ref{NK_20_15_sa_0.10} (b) and tab.~\ref{table}). Let's note that, $\beta^*$ is reached whatever the initial fitness is (except for extreme fitness values).

\paragraph{Medium temperature} 

\begin{figure}[!h] 
\begin{center} 
\includegraphics[width=0.35\textwidth]{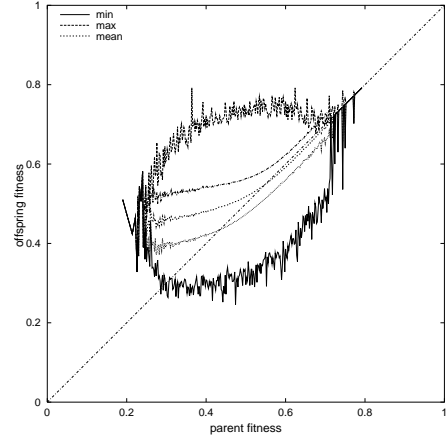} \\ 
(a) \\
\includegraphics[width=0.35\textwidth]{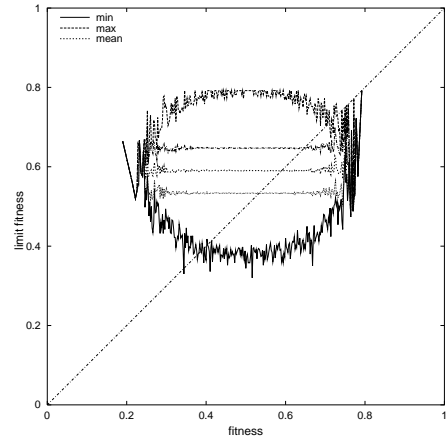} \\
(b) \\
\end{center} 
\caption{Fitness Cloud under SA at \textbf{medium temperature}: border (min and max) and mean (with standard deviation) ($FC$ (a), $FC^*$ (b)) (NK-landscape with $N=20$ and $K=15$)} 
\label{NK_20_15_sa_0.05} 
\end{figure}

At medium temperature ($T=0.05$), the fitness cloud is roughly shared by the diagonal line as the probability to accept deleterious mutation remains significant (see fig.~\ref{NK_20_15_sa_0.05} (a)). It is easy to estimate both $\beta$ and $\beta^*$ (see tab.~\ref{table}). The $\beta^*$ fitness level appears to be attractive on the limit fitness cloud (see fig.~\ref{NK_20_15_sa_0.05} (b) ~and tab.~\ref{table}). Once again, $\beta^*$ is reached whatever the initial fitness is.

\paragraph{Low temperature}

\begin{figure}[!h] 
\begin{center} 
\includegraphics[width=0.35\textwidth]{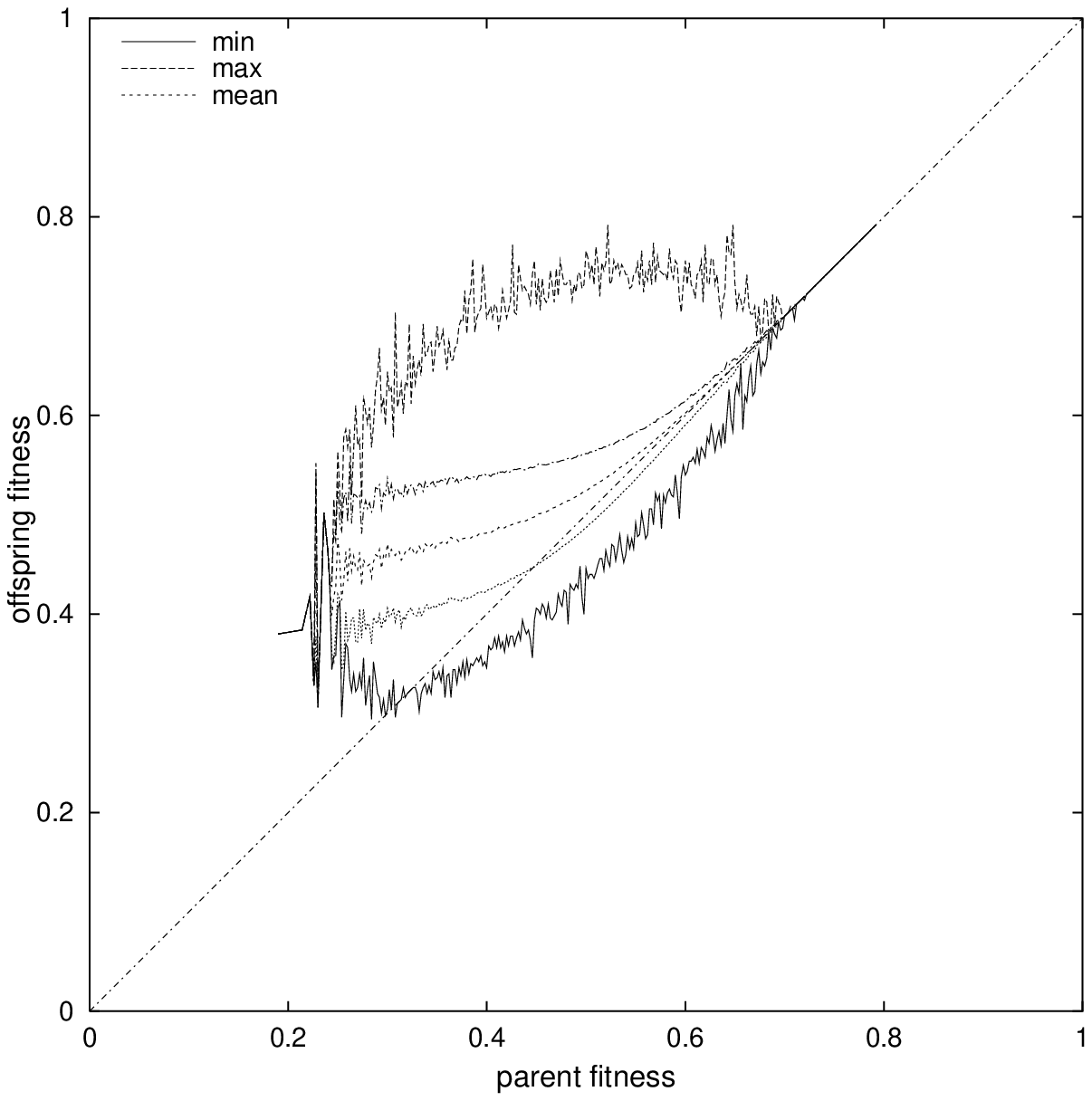} \\ 
(a) \\
\includegraphics[width=0.35\textwidth]{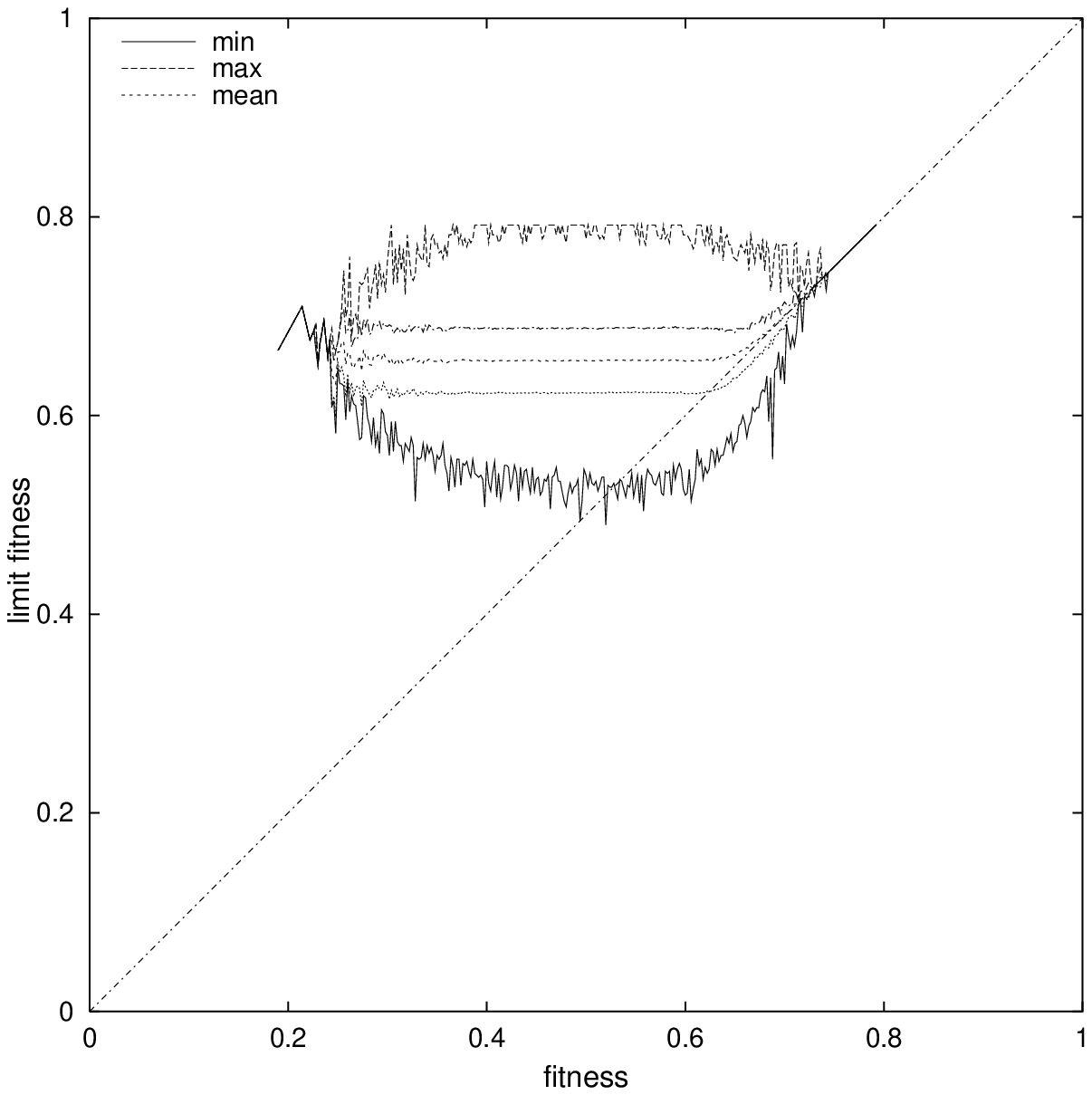} \\
(b) \\
\end{center} 
\caption{Fitness Cloud under SA at \textbf{low temperature}: border (min and max) and mean (with standard deviation) ($FC$ (a), $FC^*$ (b)) (NK-landscape with $N=20$ and $K=15$)} 
\label{NK_20_15_sa_0.01} 
\end{figure}

At low temperature ($T=0.01$), the greatest part of the fitness cloud is above the diagonal line as the probability to accept deleterious mutation is small (see fig.~\ref{NK_20_15_sa_0.01} (a)). The $FC_{mean}$ curve is a curved shape; as fitness increases, it glides slope toward the diagonal. So it is difficult to visualize point which abscissa is $\beta$. Examining data, we can find an interval where the $FC_{mean}$ curve is close to the diagonal line (with a accuracy of 0.002) (see tab.~\ref{table}). The set $FC^*_{mean}$ is roughly represented by an horizontal line except for high fitnesses where it follows the diagonal line (see fig.~\ref{NK_20_15_sa_0.01} (b) and tab.~\ref{table}). As the constant value of $f^*_{mean}$ corresponds to a bottleneck, $\beta^*$ stands for this value, although $\beta$ is not discerned.

\paragraph{Cooling process}

\begin{figure*}[!tb] 
\begin{center}
\begin{tabular}{cc} 
\includegraphics[width=0.35\textwidth]{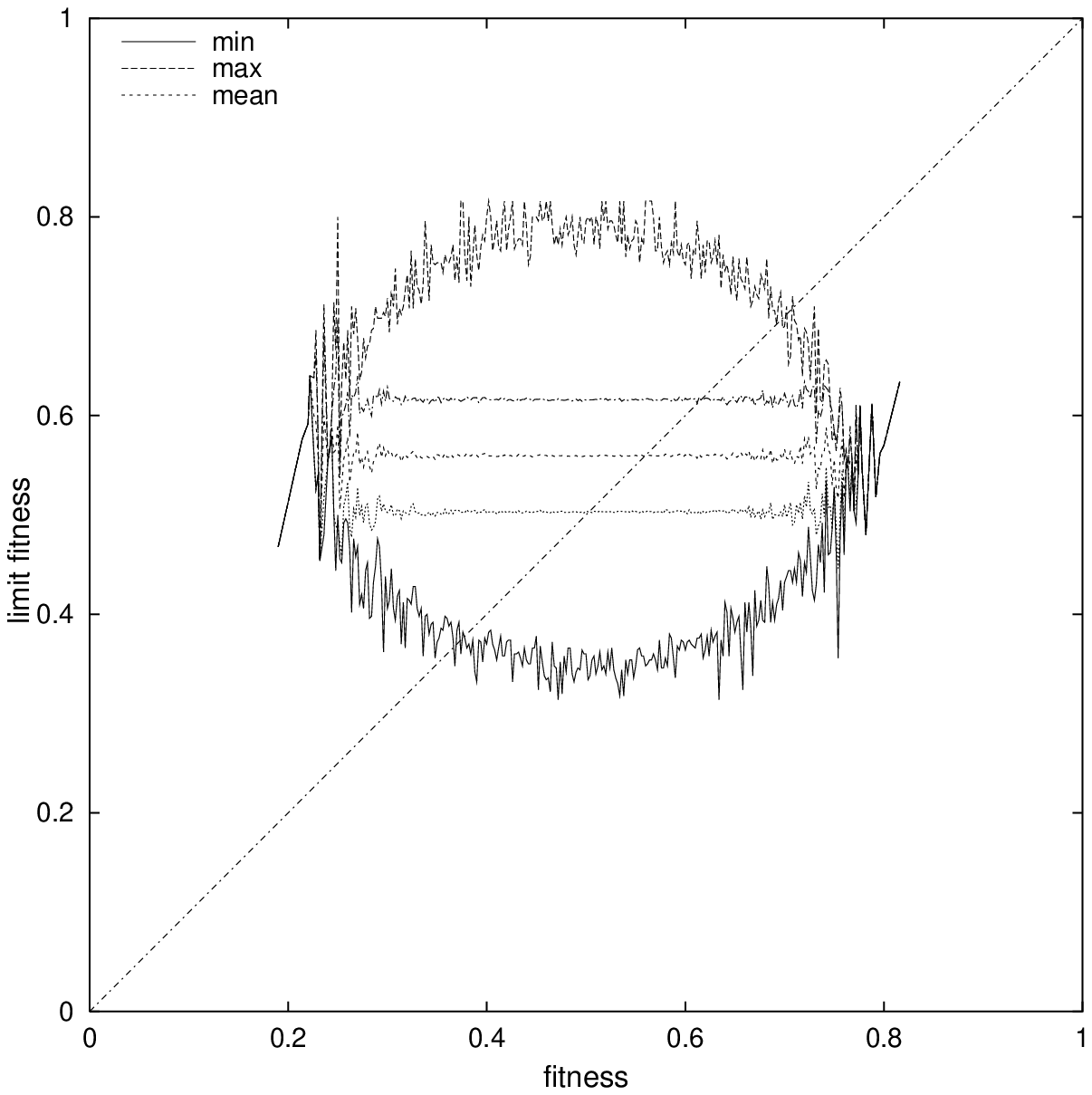} 
&
\includegraphics[width=0.35\textwidth]{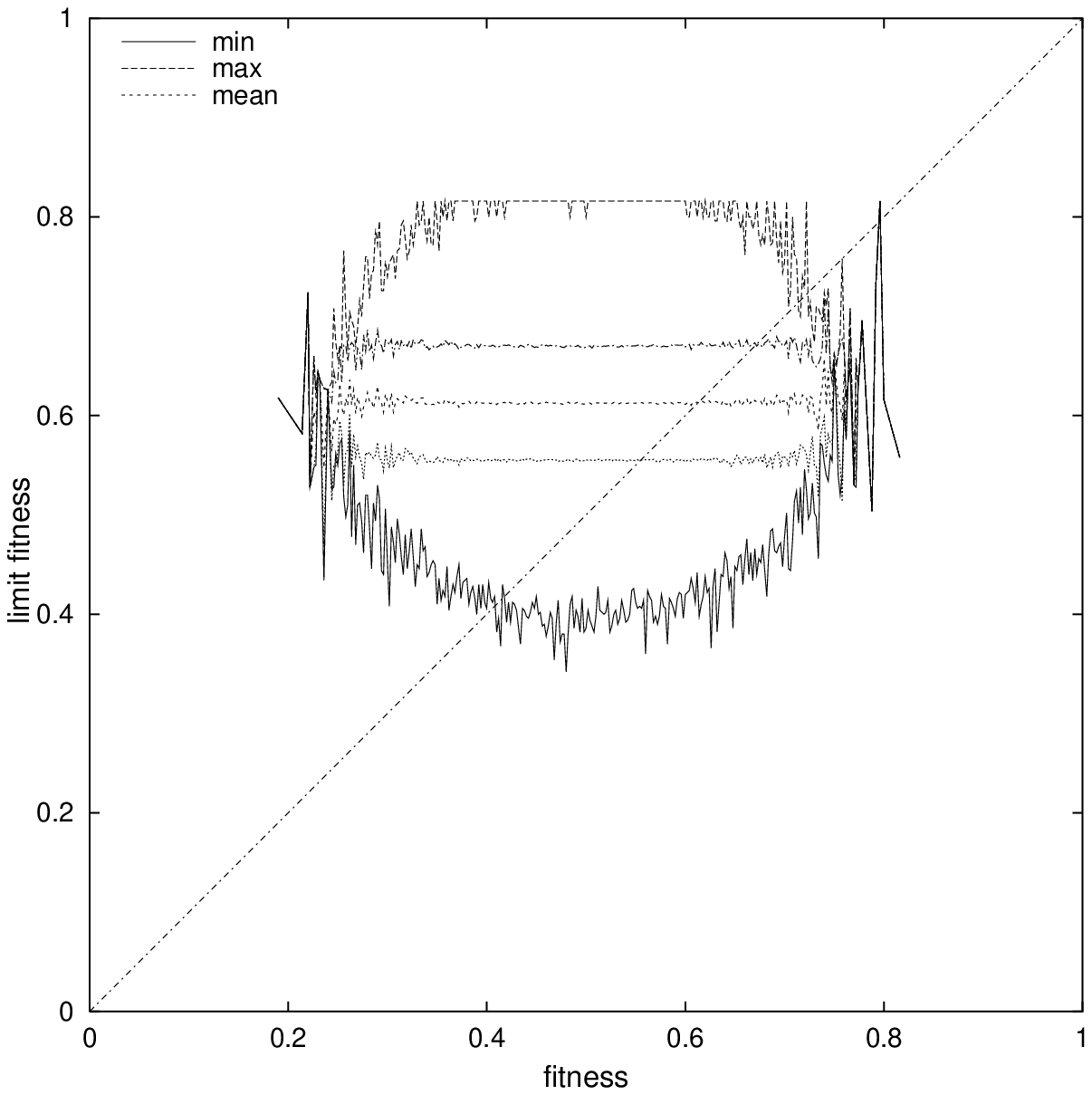} \\
(a) &
(b) \\
\includegraphics[width=0.35\textwidth]{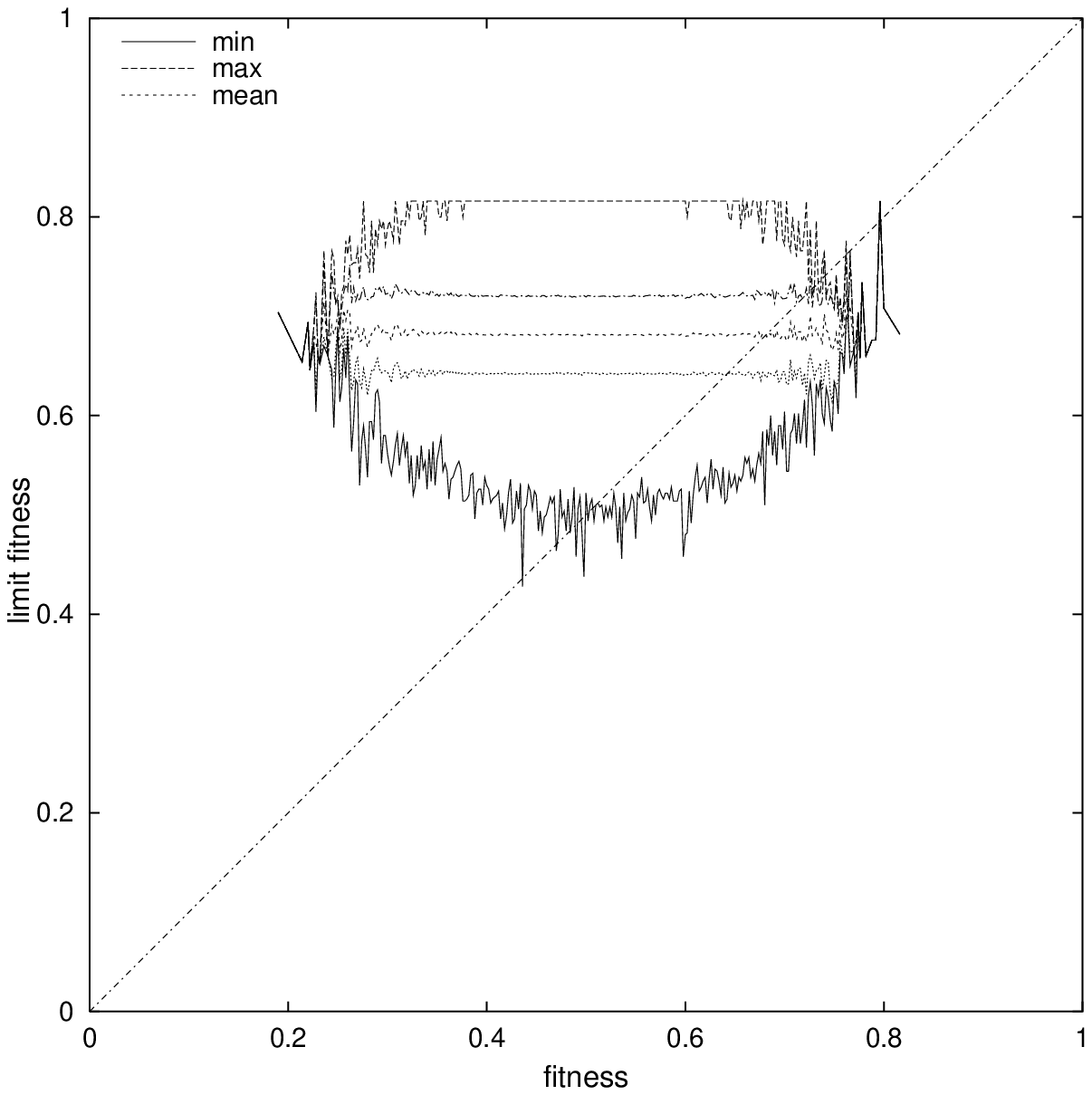}
&
\includegraphics[width=0.35\textwidth]{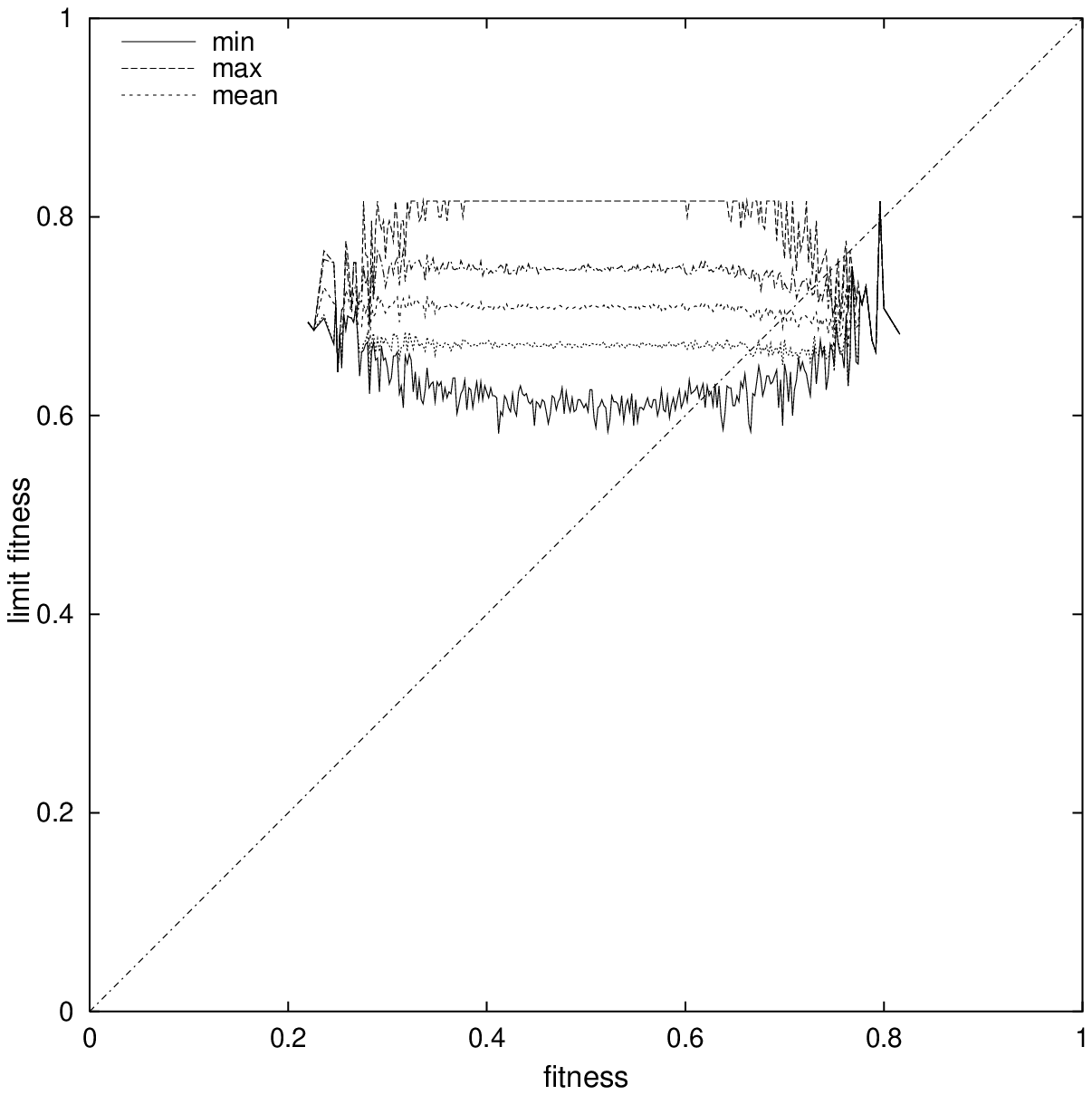} \\
(c) &
(d) \\
\end{tabular}
\end{center} 
\caption{Limit fitness cloud under SA with \textbf{cooling process}: generation 50 (a), generation 1000 (b), generation 1900 (c), generation 2450 (d) (NK-landscape with $N=20$ and $K=15$)} 
\label{NK_20_15_sa_cooling} 
\end{figure*}

The notion of system temperature is intrinsic to the SA process. By slowly lowering the temperature of an initially random system, we encourage the elements of the system to assume an orderly, minimal energy arrangement. In search process terms, a slow cooling can thus lead to an optimal state. Our annealing schedule is defined as follows: temperature starts at $0.10$, and drops to $0.01$ over a geometric decrements where $T=0.95T$\footnote{Experience has shown that the decrement coefficient should be between $0.8$ and $0.99$, with better results being found in the higher end of the range}. 2450 generations are performed; changes occurs each 50 generations. The cooling process is sufficiently slow such that for each temperature an equilibrium state, where the mean fitness no longer changes from generation to generation, is reached. Figure~\ref{NK_20_15_sa_cooling} shows snapshots of the $FC^*$ cloud at generations 50, 100, 1900 and 2450. As for low temperature, $\beta^*$ stands for the constant value of $f^*_{mean}$ since $\beta$ is not significant. During the cooling process $\beta^*$ increases with generations to finally, reaches its greatest value (see tab.~\ref{table}). Let's note that the final $\beta^*$ value for SA ($0.701$) is greater that the one for mHC ($0.667$).

\section{Fitness Cloud and neutrality : Neutral Hill Climber}
\begin{figure}[!tb] 
\begin{center} 
\includegraphics[width=0.35\textwidth]{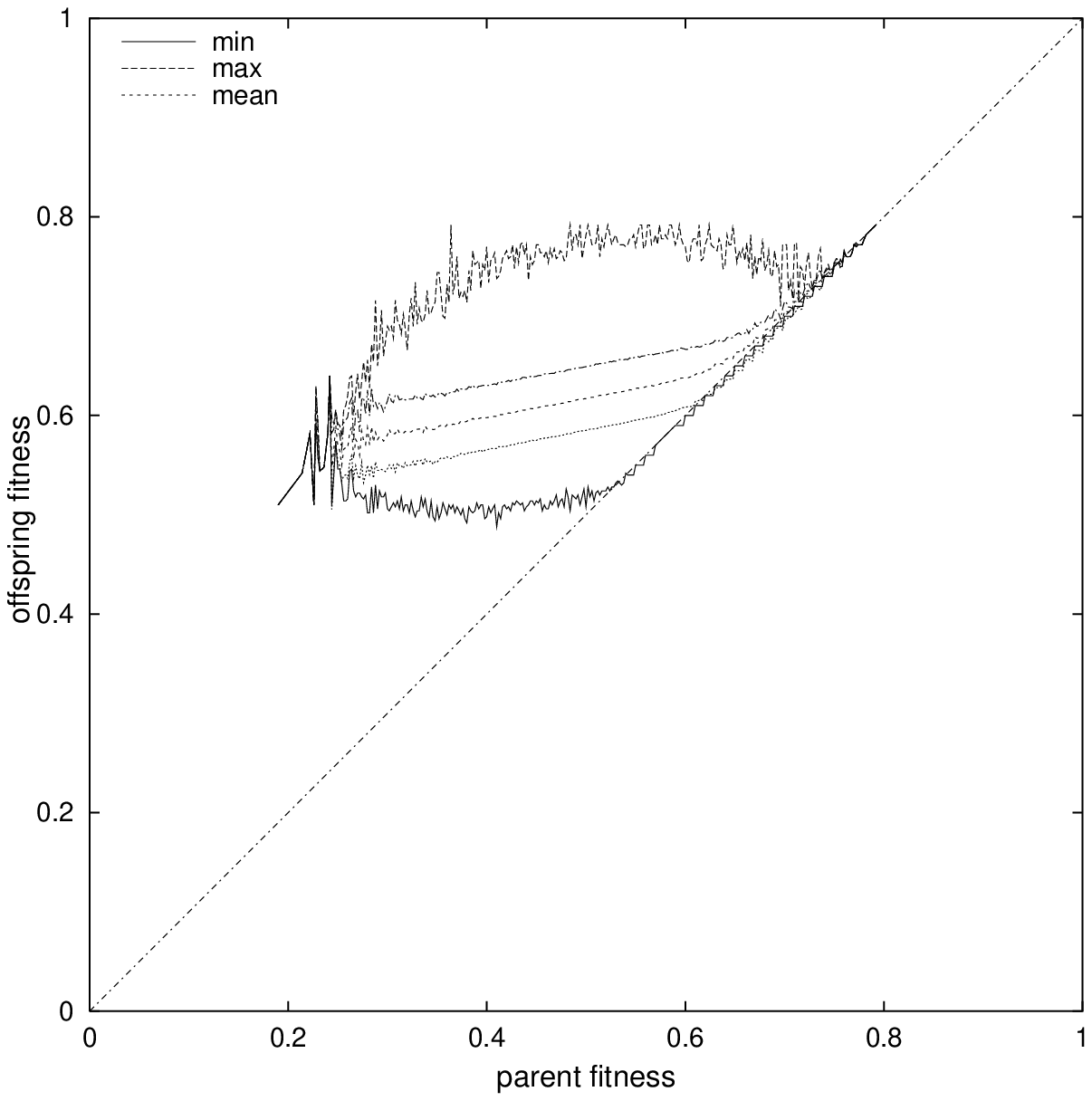} \\
(a) \\
\includegraphics[width=0.35\textwidth]{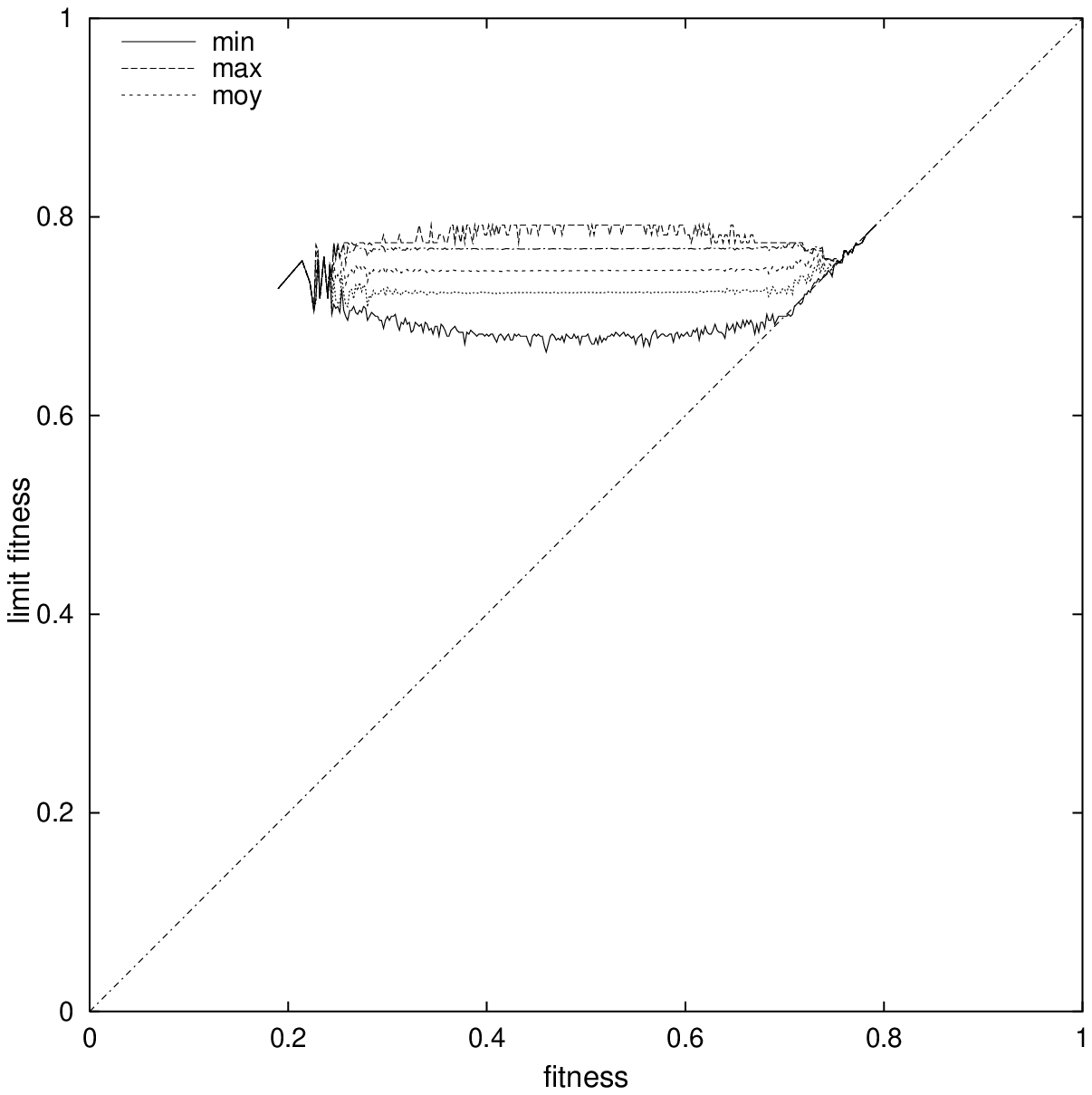} \\ 
(b) \\ 
\end{center}
\caption{Fitness Cloud under nHC: border (min and max) and mean (with standard deviation) ($FC$ (a), $FC^*$ (b)) (NK-landscape with $N=20$ and $K=15$)} 
\label{NK_20_15_sa} 
\end{figure}

In the Fitness Cloud, a vertical slice represents the set of fitnesses that could be reached from this set of neutrality. Consequently the $FC$ shows the potential interest in using neutral operator. To implement such an operator noted nOP, first the entire search space is partitioned according to fitness, then we are able to choose at random a genotype with a given fitness value\footnote{with an uniform law}. So each genotype with the same fitness are connected by an elementary neutral move. To combine neutral exploration with local search technique, we define the \textit{neutral Hill Climbing} heuristic (so-called nHC). First mHC is applied on the current genotype $g$ and the resulting change, $\Delta f = \tilde f - f$, in fitness is computed. If $\Delta f > 0$, the new point is accepted; else, we obtain offspring by applying nOP to the genotype $g$. Within a neutral set there is no productive gradient information, then the gain fitness comes from Hamming based mutation only.
The relation between $\tilde f_{mean}$, $f$, $N$ and $K$ can be derived from equation ~\ref{mean_fitness_mHC}. In this case, 
$ E(X) =  E(X(f,N,K))$ where $X(f,N,K)$ is $max( f, X_1, \ldots , X_N)$ with $X_i$ follows normal law ${\cal N} (0.5,\ \frac{1}{\sqrt{12(K+1)}})$. As a consequence, the set $FC_{mean}$ is represented by a curved shape. Figure~\ref{NK_20_15_sa} shows the $FC_{mean}$ curve is coarsely a line except for high fitness values where it glides slope toward the diagonal. So it is difficult to visualize point which abscissa is $\beta$. Examining data, we can find an interval where the $FC_{mean}$ curve is close to the diagonal line with a accuracy of 0.002 (see tab.~\ref{table}). To support the bottleneck conjecture the \textit{limit fitness cloud} is plotted (see fig. \ref{NK_20_15_sa} (b)). According to the initial fitness value we can observe two main cases about the limit behavor of the nHC heuristic. First, for an initial fitness below $\beta^*=0.746$ (as for SA with low temperature, $\beta^*$ is the constant value of $f^*_{mean}$), on average nHC converges to a fitness value close to $\beta^*$. Second, fitnesses above $\beta^*$ are fixed points in the fitness space. Therefore the $\beta^*$ fitness level is a bottleneck for the nHC heuritic. In nHC, leaving the genotype invariant instead of applying nOP, the fitness cloud remains identical. The $FC^*$ allows to show the influence of nOP: neutral exploration allows to find better fitnesses. Let's note that the fitness bottleneck for nHC ($0.746$) is greater that the one for both mHC ($0.667$) and SA ($0.701$). These experiments show the potential interest in using neutral operator when each plateau is a graph connected by a neutral operator. Of course this is an ideal case, in more realistic situations we must consider the topology graph of neutral sets induces by hamming mutation or a specific neutral operator. In real word problems one must take into account the computational cost in using neutral operator as well the availability of such operator. However in many problems neutral operator may derive from specific knowledge as symetry properties or redondancies.

\section*{Conclusion}
In this paper we have presented the Offspring-Parents Fitness Cloud. The Fitness Cloud shows evolvability as well neutrality and dynamic at fitness level of local search heuristic. We have used FC to formulate the \textit{bottleneck conjecture}: there is a fitness level in the NK-landscape that causes local search process to slow down and stop. Moreover the FC is useful in identifying the bottleneck's location in the neighbourhood of a fitness value $\beta$. This conjecture deals with the average behavior of local search heuristics only. To confirm this conjecture, the Limit Fitness Cloud is introduced; it gives a visual rendering of the long term behavior. The experiments performed on NK-landscape with Hill Climbing, Simulated Annealing and Neutral Hill Climbing support our conjecture. Indeed there is a fitness level $\beta^*$ close to $\beta$ which is difficult to bypass. FC can also represent the neutrality of landscape. So, we have designed the Neutral Hill Climbing (nHC) which shows how to exploit the information about neutrality given by FC. We have found experimentaly that the bottleneck for nHC is greater that the one for both mHC and SA.\\
\commentaire{
the \textit{fitness cloud} as a complementary viewpoint to the \textit{fitness landscape} metaphor. FC is a 2-d representation where topology induced by a heuristic is directly taken into account. Our analytical and empirical results suggest that it is possible to characterize the set of local optima and barriers of fitness too. Despite the fact that one motivation for the landscape metaphor is to think of a heuristic search as a way to travel across the landscape, getting a geographical view on the search space gives little information to predict the performance. In such a context, we believe the FC can be utilized beneficially to track the dynamic and to predict the average behaviour of the search process. Changing the metaphor from landscape to cloud, allow to swap from point getting stuck in local optimum to point pulled towards a particular set of neutrality.}
In this work exhaustive enumeration of the search space is used; future works should address the question of how to get FC from data collected through random sampling or during the search process. To track the dynamics of population based heuristics, as \textit{genetic algorithms}, notion of population must be taken into account. The approach can be extended to others operators than local variations, in particular we project to draw the FC for crossover. Another extension is to study the effect of choosing one representation: for instance, what happen to the FC when switching from Integer coding to Gray coding ? The NK-landscapes can be useful for initial investigations, but results gained on them cannot be guaranteed to transfer to real word problems in general. So we obviously have to study fitness cloud and the fitness bottleneck hypothesis on other problems than NK-landscapes.


\bibliography{Biblio}

\begin{thebibliography}{10}

\bibitem{ALT:94}
Lee Altenberg.
\newblock The evolution of evolvability in genetic programming.
\newblock In {\em In Kinnear, Kim (editor). Advances in Genetic Programming.
  Cambrige, MA}, pages 47--74. The MIT Press, 1994.

\bibitem{WA-AL}
L.~Altenberg G.~P.~Wagner.
\newblock Complexes adaptations and the evolution of evolvability.
\newblock In {\em Evolution}, pages 967--976, 1996.

\bibitem{GRE:95}
J.~J. Grefenstette.
\newblock Predictive models using fitness distributions of genetic operators.
\newblock In D.~Whitley, editor, {\em Foundations of Genetic Algorithms}, San
  Mateo, CA, 1995. Morgan Kaufmann Publishers.

\bibitem{KAU:93}
S.~A. Kauffman.
\newblock {\em ``The origins of order''. {Self}-organization and selection in
  evolution}.
\newblock Oxford University Press, New-York, 1993.

\bibitem{KIM:83}
M.~Kimura.
\newblock {\em The Neutral Theory of Molecular Evolution}.
\newblock Cambridge University Press, Cambridge, UK, 1983.

\bibitem{KIR:83}
S.~Kirkpatrick, C.~D.~Gelatt Jr., and M.~P. Vecchi.
\newblock Optimization by simulated annealing.
\newblock In {\em Science}, pages 671--680, 1983.

\bibitem{MAN:91}
B.~Manderick, M.~de~Weger, and P.~Spiessens.
\newblock The genetic algorithm and the structure of the fitness landscape.
\newblock {\em Proceedings of the Fourth International Conference on Genetic
  Algorithms}, pages 143--150, 1991.

\bibitem{ROS:96}
Helge Ros\'e, Werner Ebeling, and Torsten Asselmeyer.
\newblock The density of states - a measure of the difficulty of optimisation
  problems.
\newblock In {\em Parallel Problem Solving from Nature}, pages 208--217, 1996.

\bibitem{SMI:01}
Smith, Husbands, Layzell, and O'Shea.
\newblock Fitness landscapes and evolvability.
\newblock {\em Evolutionary Computation}, 1(10):1--34, 2001.

\bibitem{WEI:90}
E.~D. Weinberger.
\newblock Correlated and uncorrelatated fitness landscapes and how to tell the
  difference.
\newblock In {\em Biological Cybernetics}, pages 63:325--336, 1990.

\bibitem{WEI:96}
E.~D. Weinberger.
\newblock {NP} completeness of kauffman's {NK} model, a tuneable rugged fitness
  landscape.
\newblock {\em Santa Fe Institute: Working Papers 96-02-003}, 1996.

\bibitem{WRI:69}
S.~Wright.
\newblock The theory of gene frequency.
\newblock In {\em Evolution and the genetics of Population}, volume~2, pages
  120--143. University of Chicago Press, 1969.

\end{thebibliography}
\bibliographystyle{plain} 

\end{document}